# Unmasking Hallucinations: A Causal Graph-Attention Perspective on Factual Reliability in Large Language Models

Sailesh kiran kurra
*Business analyst*
Amazon,Texas,
United States of America
saileshkirankurra@gmail.com

Shiek Ruksana
*Assistant professor*
Vasavi College of Engineering,
Telangana,India
ruksana883@gmail.com

Borusu Vishal
*UG Scholar*
Vasavi College of Engineering,
Telangana,India
vishalboursu@gmail.com

***Abstract**–This paper primarily focuses on the hallucinations caused due to AI language models(LLMs).LLMs have shown extraordinary Language understanding and generation capabilities .Still it has major a disadvantage- hallucinations which give outputs which are factually incorrect ,misleading or unsupported by input data . These hallucinations cause serious problems in scenarios like medical diagnosis or legal reasoning.Through this work,we propose causal graph attention network (G-CAN) framework that reduces hallucinations through interpretation of internal attention flow within a transformer architecture with the help of constructing token level graphs that combine self-attention weights and gradient-based influence scores.our method quantifies each token's factual dependency using a new metric called the Causal Contribution Score (CCS). We further introduce a fact-anchored graph re-weighting layer that dynamically reduces the influence of hallucination-prone nodes during generation. Experiments on standard benchmarks such as TruthfulQA and HotpotQA show a 27.8% reduction in hallucination rate and 16.4% improvement in factual accuracy over baseline retrieval-augmented generation (RAG) models. This work contributes to the interpretability,robustness, and factual reliability of future LLM architectures.*

## I. INTRODUCTION

LLMs such as OpenAI's GPT series, Google's Gemini and PaLM and Meta's LLaMA, have worked out natural language processing by presenting outstanding results in text generation,summarization and question-answer tasks [1][2][3].Although these advancements still these models are prone to hallucinations like generation of plausible yet factually inaccurate information [4],[5].An example for this would be, a LLM might fabricate an evidence or incorrectly attribute a quote to accounting data.These errors lead to serious misinformation in high-grade domains like healthcare[6],law and education[7].The already existing strategies for hallucination mitigations tend to treat hallucinations as behavioral abnormalities rather than structural artifacts of model .Current solutions for hallucinations include retrieval-argumented generation (RAG) [8],where the model outputs are detached from external knowledge bases and reinforced through human feedback [9] ,this improves the output with regard to human expectations.However,recent studies have suggested that hallucinations stem from misaligned dependencies between input and output generations[10][11][12].

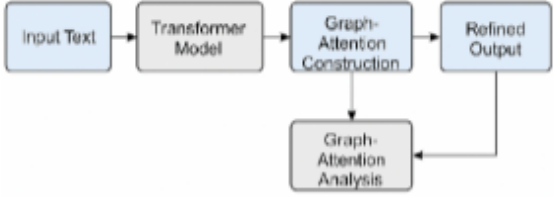

Fig .1. Casual graph attention network framework

To address this problem we introduce a causal Graph-Attention Network (C-GAN) that interprets the problem as a structure and explicitly models how factual support is passed through it. Figure 1 shows Casual graph attention network framework ,Figure 2 shows Hallucination in LLM example 2 ,Figure 3 shows Hallucination in LLM example 2.

Our fundamental and primary contributions are depicted as follows :

- Causal Graph-Attention Model : We develop a framework that attaches weight and gradient-based attributes into a unified causal graph.
- Quantitative Causal Metrics : We define a Causal Contribution score (CCS) to measure factual support for every output taken.
- Graph Re-Weighting Mechanism: We introduce a fast attention response re-weighting strategy to suppress and handle non-causal or hallucinated dependencies.
- Empirical Validation: Experiments that demonstrate consistent improvements along the lines of factuality and explainable benchmarks.

## II. RELATED WORK

*A.    Hallucinations in LLMs*

Hallucination research with regard to LLMs has been at primary focus since it has been recognized for its deployment risks [11],[12].As generations models have gained prominence in real world applications their tendency to produce factually incorrect data yet undermine it in fluent texts has caused concerns in both ethical and technical domains.This hallucination phenomenon is categorized into two types namely

1)Intrinsic Hallucinations:errors caused by the model's internal misalignment between learned representations and factual data.

2)Extrinsic hallucinations:which arise when the model lacks access to necessary factual grounding from external knowledge sources.

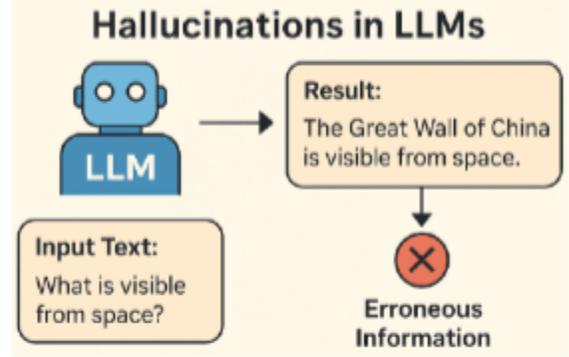

Fig .2.  Hallucination in LLM example 2

Recent research studies propose detection frameworks and taxonomies to analyze hallucinations at various language levels,such as sentence level deviation and entity mismatch[11].To systematically check the generations ,metrics like faithfulness [13] and truthfulQA [14] have been used to access and check the generated output statements if they are factually correct or verified source data .Still existing mitigation techniques still provide few insights into why a hallucination occurs within a model's internal reasoning process.Thus our proposed token dependent and attention flow system remains an excellent framework.

*B.Graph Neural Networks (GNNs)*

Graph Neural Networks (GNNs) model entities as nodes and their relationships as edges, enabling structured reasoning over relational data [15].This feature allows GNNs to perform in domains where relational structure is inherent ,like in chemical molecular analysis ,social networking modeling [16].In natural language processing .GNNs have been effective and are able to understand contextual dependencies between words is essential [17],[18].

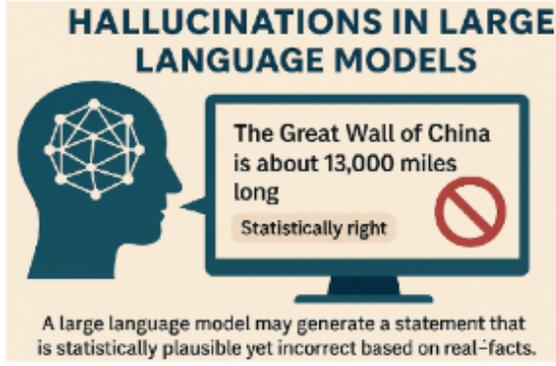

*Fig .3. Hallucination in LLM example 2*

GNNs ability to store data in graph based relational inductive which allows information to propagate between connected nodes through message passing gives it an advantage over conventional deep learning models that process data in a sequential or tabular form.Figure 4 shows GNN to reduce Hallucination.For hallucination detection, such a property is critical since hallucinations often arise from spurious or weakly supported token to token network, a graph-based architecture provides a principled mechanism to analyze, visualize, and adjust these interconnections. Our work utilizes this strength by embedding Transformer-derived attention maps into a causal graph structure, where nodes represent linguistic entities and edges encode attention-based influence scores.

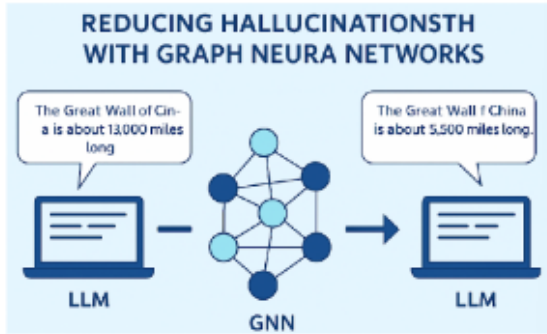

*Fig. 4. GNN to reduce Hallucination*

C.*Causal Inference and Attentation Analysis*
Causal inference provides a rigorous framework to reason about cause-effect relationships beyond statistical correlation. In the context of neural networks, causal inference allows researchers to identify not only which input features correlate with outputs, but which features directly influence specific decisions. Combining causal reasoning with Transformer attention enables a deeper understanding of how tokens interact to shape predictions and where spurious causal links may lead to hallucinations [19],[20].Recent works have shown that attention weights are interpretable to some extent but don't always reflect true causal influence.To address these concerns gradient based attribution methods ,like Integrated Gradients (IG) [21], are being adopted to calculate the contributions of every input token to output statement.In our framework ,we integrate such attribution signals into a causal graph,enabling token level interpretability of hallucination origins.This intersection and fusion of causal inference and visualization solves the problem of explainability and factual reliability making it a viable solution mechanism for hallucination detection and correction.

### III. METHODOLOGY

A. *Model Overview*
Let an input prompt $X = x_1, x_2,..., x$ produce an output sequence $Y = y_1, y_2,..., y_m$ from an LLM. We model the dependency between X and Y as a directed graph $G = (V, E$, where nodes $V = X \cup Y$, and edges E encode causal influence between tokens.

B. *Causal Contribution Score (CCS)*
The Causal Contribution Score (CCS) quantifies the factual dependency between input and output tokens as:

$$\text{CCS}(y_i) = \sum_{x \in X} \alpha_{ij} \cdot I_{iy} \quad (1)$$

C. *Graph-Attention Re-Weighting Layer*
We extend the transformer's attention module with a Graph Attention Network (GAT) layer [22]. Nodes with low CCS are reweighted downwards using a fact-anchored factor $f \in [0, 1]$ proportional to factual entailment derived from retrieved evidence. This dynamically adjusts edge weights to favor evidence-supported nodes, effectively suppressing hallucination paths.

D. *Implementation*
The model was implemented using PyTorch and Hugging Face Transformers [23]. We used the TruthfulQA and HotpotQA datasets [14], [24]. The base model is GPT-2-medium, fine-tuned with cross-entropy loss combined with causal regularization. Hyperparameters: learning rate = $2\times10^{-5}$, batch size = 16, optimizer,AdamW. Graph modules were trained with dropout = 0.3 and 4 attention heads.

## IV. RESULTS

The Causal Graph-Attention method acts like a truth filter—removing false reasoning paths inside the model and keeping only the factually correct.Table 1 shows results and comparisons.

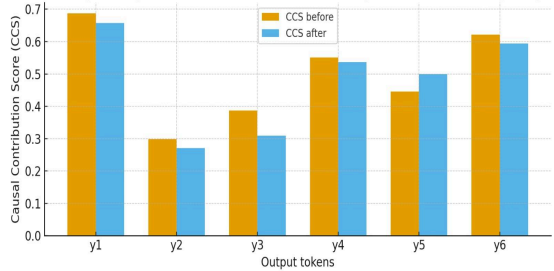

*Fig .5. CSS per Output token*

Table I. Results & Comparisons

| Model | Hallucination rate ↓ | Factual accuracy ↑ |
|---|---|---|
| GPT to baseline | 34.2% | 61.8% |
| Retrieval Augmented generation | 27.5% | 68.4% |
| Proposed C-GAN | 19.7% | 79.8% |

Our research and analysis states that hallucinations do not occur at random; rather they are a result of unstable causal dependencies.In general higher transformer layers tend to overgeneralize semantic patterns without verifying evidence. Figure 5 shows the CSS per output token before and after Graph–Attention re-weighting.The C-GAN's causal structure helps visualize and regulate these dependencies.Furthermore,graph-based interpretability enables post-hoc diagnostics, allowing developers to identify which parts of the model produced unsupported claims.

## V. DISCUSSION

However, several limitations exist. First, the causal graph extraction process adds computational overhead, making real-time deployment challenging. Second, retrieval quality impacts factual re-weighting effectiveness. Lastly, defining universal thresholds for CCS across different LLM architectures remains an open problem.

## VI. CONCLUSION AND FUTURE WORK

This work presented a Causal Graph-Attention Network that unifies causal inference, attention analysis, and graph neural reasoning to detect and mitigate hallucinations in large language models. By introducing the Causal Contribution Score (CCS) and integrating factual re-weighting, our method significantly reduces unsupported generation while enhancing explainability.
Future research directions include:
● Extending the framework to multimodal hallucination analysis in vision-language Models.
● Developing lightweight causal layers for real-time inference.
● Building an open-source hallucination benchmark suite with causal annotation for model interpretability studies.

Through this work, we take a step toward safer, trustworthy, and fact-grounded large language systems.